\documentclass{article} 
\usepackage{iclr2027_conference,times}

\usepackage{amsmath,amsfonts,bm}

\def\eqref#1{equation~\ref{#1}}

\def\1{\bm{1}}

\DeclareMathAlphabet{\mathsfit}{\encodingdefault}{\sfdefault}{m}{sl}
\SetMathAlphabet{\mathsfit}{bold}{\encodingdefault}{\sfdefault}{bx}{n}

\usepackage{hyperref}
\usepackage{url}

\usepackage{amsmath, amssymb, amsfonts, amsthm}
\usepackage{paralist}
\usepackage{booktabs}
\usepackage{graphicx}
\usepackage{multirow}
\usepackage{wrapfig}
\usepackage[table]{xcolor}
\usepackage[most]{tcolorbox}
\usepackage{enumitem}
\definecolor{insightpurple}{RGB}{138,43,226}

\fancypagestyle{firstpage}{
    \fancyhead{}
    \fancyhead[L]{\includegraphics[width=1.0in]{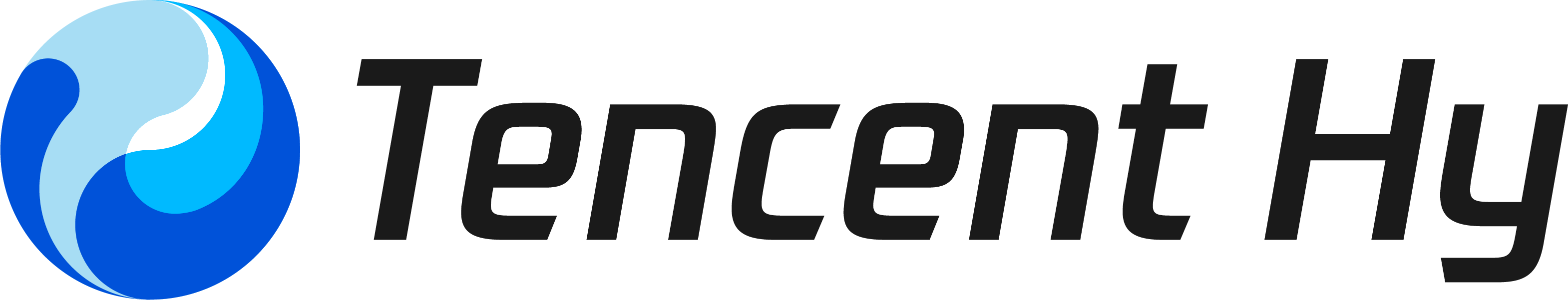}}
}

\makeatletter
\newcommand{\fixedthanks}[2]{%
    \footnotemark[#1]%
    \protected@xdef\@thanks{%
        \@thanks
        \protect\footnotetext[#1]{#2}}%
}
\makeatother

\newtcolorbox{contributionbox}{
    enhanced,
    colback=insightpurple!5,
    colframe=insightpurple!50!black!50!,
    boxrule=0.5mm,
    arc=2mm,
    left=4pt,
    right=8pt,
    top=0pt,
    bottom=0pt
}

\tcbset{
    statementbox/.style={
        enhanced,
        breakable,
        colback=cyan!5!white,
        colframe=cyan!45!blue!60,
        colbacktitle=cyan!45!blue!60,
        coltitle=white,
        fonttitle=\bfseries,
        arc=2mm,
        boxrule=0.8pt,
        left=8pt,
        right=8pt,
        top=6pt,
        bottom=6pt
    }
}

\newcounter{finding}
\newcounter{prompt}

\newenvironment{finding}
{%
    \refstepcounter{finding}%
    \begin{tcolorbox}[
        statementbox,
        title={Finding \thefinding}
    ]%
}
{%
    \end{tcolorbox}%
}

\newenvironment{prompt}[1]
{%
    \refstepcounter{prompt}%
    \begin{tcolorbox}[
        statementbox,
        title={Prompt \theprompt: #1}
    ]%
}
{%
    \end{tcolorbox}%
}

\definecolor{cmath}{HTML}{E4635C}
\definecolor{cscience}{HTML}{4CB391}
\definecolor{ccode}{HTML}{3B5B92}
\definecolor{cif}{HTML}{E8A33D}
\definecolor{cagent}{HTML}{8E6BAB}

\usepackage{xfp}
\usepackage{siunitx}
\newcommand{\mk}[1]{#1}
\newcommand{\mkstar}[1]{#1\textsuperscript{*}}
\newcommand{\bd}[1]{#1}
\newcommand{\bdbold}[1]{\textbf{#1}}
\newcommand{\cellmarks}{\let\mk\mkstar\let\bd\bdbold}
\newcommand{\rowavg}[9][]{%
  {\cellmarks #2} & {\cellmarks #3} & {\cellmarks #4} &
  {\cellmarks #5} & {\cellmarks #6} & {\cellmarks #7} &
  {\cellmarks #8} & {\cellmarks #9} &
  #1{\num[mode=text,reset-text-series=false,
          round-mode=places,round-precision=1,round-pad=true]%
        {\fpeval{(#2+#3+#4+#5+#6+#7+#8+#9)/8}}}}

\title{Consolidating RLVR Capabilities Across \\ Domains: A Deep Dive into Fusion Paradigms
\vspace{-0.3em}
}

\author{
Siye Wu\textsuperscript{\(\spadesuit\clubsuit\)}\fixedthanks{2}{Work done during internships at Tencent. Contact: Siye Wu \href{mailto:siyewu24@m.fudan.edu.cn}{\textless{}siyewu24@m.fudan.edu.cn\textgreater{}}.},\enspace
Kai Yang\textsuperscript{\(\clubsuit\)},\enspace
Yuchen Cai\textsuperscript{\(\clubsuit\)}\footnotemark[2],\enspace
Xin Xu\textsuperscript{\(\clubsuit\)},\enspace
Peng-Yuan Wang\textsuperscript{\(\clubsuit\)}\footnotemark[2],\enspace
Jiaxuan Wang\textsuperscript{\(\clubsuit\)}\footnotemark[2],\enspace \\
Jiashun Liu\textsuperscript{\(\clubsuit\)}\footnotemark[2],\enspace
Jiafei Lyu\textsuperscript{\(\clubsuit\)},\enspace
Yangkun Chen\textsuperscript{\(\clubsuit\)},\enspace
Saiyong Yang\textsuperscript{\(\clubsuit\)},\enspace
Yanghua Xiao\textsuperscript{\(\spadesuit\)}\fixedthanks{1}{Corresponding author: \href{mailto:shawyh@fudan.edu.cn}{\textless{}shawyh@fudan.edu.cn\textgreater{}}.} \\
\normalfont
\textsuperscript{\(\spadesuit\)}Fudan University \quad
\textsuperscript{\(\clubsuit\)}LLM Department, Tencent\\
\vspace{0.2em}
\normalfont{
\href{https://github.com/Di-viner/LLM-Fusion}{\raisebox{-0.15em}{\includegraphics[height=1.1em]{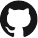}}\,GitHub}
\quad \quad
\href{https://huggingface.co/collections/Siye01/llm-fusion}{\raisebox{-0.15em}{\includegraphics[height=1.1em]{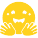}}\,Hugging Face}
}
}

\iclrfinalcopy 
\begin{document}

\vspace*{-1.6em}
\maketitle
\fancyhead{}
\thispagestyle{firstpage}

\vspace*{-0.4em}
\begin{abstract}
Reinforcement learning with verifiable rewards (RLVR) improves specific capabilities of large language models, but covering multiple capabilities often involves training separate domain experts and subsequently consolidating them.
We organize three fusion paradigms by the artifacts they reuse: \textbf{Merge} combines expert task vectors, \textbf{Mix RL} pools their datasets, and multi-teacher on-policy distillation (\textbf{MOPD}) uses both.
Because they have largely been studied in isolation, how they compare and how to choose among them remain unclear.
We compare all three using shared experts and data across model scales and a multi-domain benchmark suite.
Although their average performance differs by at most \(1.4\) points, the gap reaches \(8.6\) points on a single benchmark, with domain-level variation tracking cross-domain relations visible in task-vector geometry.
Training dynamics expose distinct constraints: Mix RL depends on domain mixture proportions, MOPD remains bounded by its teachers, and Merge compresses all expert updates into one.
All three improve single-sample accuracy without measurable gains in solution coverage or losses in held-out capabilities. 
These results yield a practical guideline: use Merge when experts already exist and cheap fusion is paramount; Mix RL when training a unified model without experts, with domain proportions adjusted for cross-domain transfer; and MOPD when preserving domain-specific gains matters more than surpassing teachers or minimizing end-to-end cost.
\end{abstract}

\section{Introduction}
\label{sec: introduction}
Reinforcement learning with verifiable rewards (RLVR) has become a standard approach to post-training large language models (LLMs)~\citep{jaech2024openai, guo2025deepseek}.
In RLVR, a verifier scores responses sampled from the model, and an RL algorithm---often a group-based method such as Group Relative Policy Optimization (GRPO)~\citep{shao2024deepseekmath}---turns those scores into policy updates~\citep{guo2025deepseek}.
In practice, RLVR is typically used to improve performance in a specific domain, such as math~\citep{hu2025openreasonerzero} or code~\citep{feng2026retool}, with rewards tailored to that domain.
Extending these gains across multiple domains typically requires separate training runs, producing one expert per domain~\citep{blakeman2025nemotron}.
The growing demand for generalist models therefore calls for consolidating these separately trained domain experts into a single model.

\begin{figure*}[t]
    \centering
    \includegraphics[width=\linewidth]{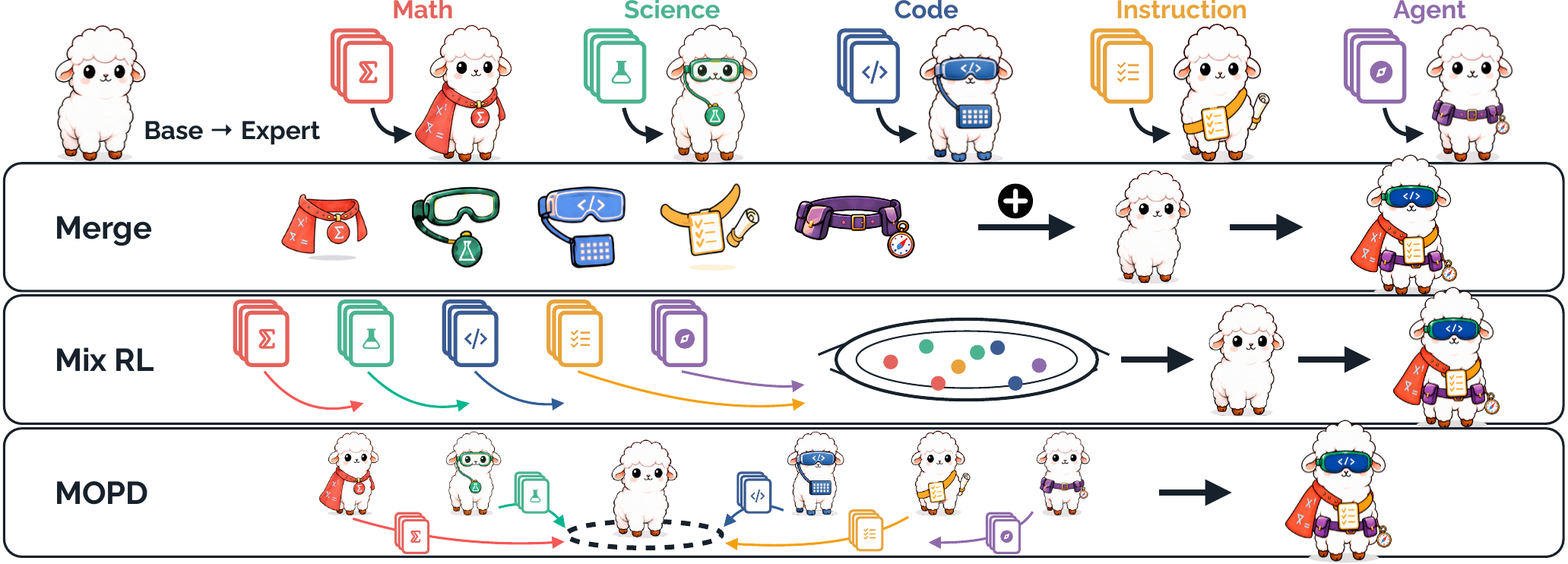}
    \vspace{-1.6em}
    \caption{Three paradigms for consolidating domain-specific capabilities acquired through RLVR into one model. Merge combines expert task vectors, Mix RL trains jointly on mixed-domain data, and MOPD draws on both through multi-teacher on-policy distillation.}
    \label{fig:intro}
    \vspace{-1.1em}
\end{figure*}

As illustrated in Figure~\ref{fig:intro}, we organize these approaches into three paradigms according to the artifacts they reuse.
Each expert \(i\) is associated with two artifacts that can be reused for fusion: its \textit{task vector} \(\tau_i = \theta_i - \theta_0\), which records its displacement from the base weights, and the \textit{dataset} \(\mathcal{D}_i\) on which it was trained.
\textbf{Merge} keeps the \(\tau_i\) and folds them back into the base weights, with no training of its own~\citep{ilharco2023editing}.
\textbf{Mix RL} sets the experts aside and keeps the \(\mathcal{D}_i\), sampling from all of them in a single RLVR run~\citep{huang2026step}.
Multi-teacher on-policy distillation (\textbf{MOPD}) keeps both, supervising a student on its own rollouts with logits from the expert corresponding to each prompt's domain~\citep{lu2025onpolicydistillation}.
By reusing different artifacts, the paradigms differ in prerequisites, costs, and supervision, underscoring the need for a controlled comparison.

However, how the three paradigms compare with one another remains unclear.
This uncertainty persists since the paradigms are typically evaluated in isolation on different backbones and benchmarks, hindering direct comparison and obscuring the source of domain-level gains and losses.
Recent studies have compared mixed RLVR with post-hoc fusion~\citep{wang2026mix, gu2026co}, but none has evaluated all three under a common framework.
In this paper, we systematically compare the three paradigms in a unified setting, asking what each achieves and why their outcomes differ.
To this end, we train full-parameter and LoRA experts for five domains on each of the 4B and 8B Qwen3 backbones~\citep{yang2025qwen3} and evaluate all three across a diverse set of benchmarks.

Our comparison shows that aggregate performance can mask important domain-level differences.
Their average performance differs by at most \(1.4\) points on either backbone, yet the gap reaches \(8.6\) on a single benchmark.
These systematic differences make the paradigms non-interchangeable and reflect relations among the fused domains.
We characterize these relations through expert behavior and task-vector geometry.
Both reveal the same structure: math, science and code reinforce one another, whereas instruction following and agentic tool use are nearly orthogonal to the other domains.
The resulting domain-level outcomes reflect how each paradigm allocates or preserves domain-specific learning signals.
Where cross-domain transfer is absent, Mix RL progresses on a domain in proportion to the prompts sampled from it.
MOPD preserves each domain through token-level teacher supervision but remains bounded by its teachers.
Merge compresses five task vectors into a single update, so the fraction of each expert's gain that survives varies widely across domains.

Two further measurements bound what fusion ultimately delivers.
All three paradigms improve over the base model at \(\mathrm{pass@}1\), but their advantage shrinks with additional samples; at \(\mathrm{pass@}32\), all three are indistinguishable from the base model on AIME.
Fusion therefore improves single-sample accuracy without expanding the set of solvable problems, while preserving the base model's existing capabilities.
Consistent with this preservation, no paradigm underperforms the base model on either factual recall or long-context reasoning, both of which are held out from training.
The paradigms also differ markedly in their prerequisites, and their compute costs span more than an order of magnitude.
Once the experts exist, Merge requires only minutes of arithmetic.
Mix RL needs no experts and costs about \(0.6\times\) the per-domain RL reference, but depends on careful data mixing.
MOPD has the cheapest trained fusion stage, below \(0.2\times\) the reference, yet the highest end-to-end cost because it presupposes the experts it distills.

\begin{contributionbox}
    Our contributions are as follows.
    \begin{enumerate}[leftmargin=1.5em]
        \item A controlled comparison of Merge, Mix RL and MOPD under a unified experimental framework, organized by the artifacts they reuse and evaluated across model scales using shared experts, training data and a diverse multi-domain benchmark suite.
        \item An explanation of their domain-level outcomes across all five target domains through cross-domain relations in behavior and task-vector geometry, together with analyses of training steps, sampled prompts and supervision throughout the training process.
        \item A characterization of fusion's benefits and limits, showing no broader solution coverage or measurable held-out capability loss, and providing guidance for choosing among paradigms based on domain relations, costs and prerequisites.
    \end{enumerate}
\end{contributionbox}

\section{Related Work}
\label{sec: related}
\paragraph{Multi-domain RLVR.}
Recently, reinforcement learning with verifiable rewards (RLVR) has attracted substantial attention for its effectiveness in enhancing LLM capabilities~\citep{lamberttulu, guo2025deepseek}.
Using simple rule-based rewards, group-based RLVR algorithms such as Group Relative Policy Optimization (GRPO)~\citep{shao2024deepseekmath} enable models to achieve expert-level performance in specific domains~\citep{hu2025openreasonerzero, feng2026retool}.
However, such gains are typically domain-specific, and extending expert-level performance across multiple domains remains non-trivial because domains may reinforce or interfere with one another~\citep{li2025can}.
Related phenomena have been investigated in multi-task learning, where cross-domain interactions are commonly characterized through task relatedness, representation sharing, and optimization-level interference~\citep{yu2020gradient, wu2026imbalanced}.
Prior studies have primarily focused on specific reasoning domains and final performance metrics.
In contrast, we consider a broader set of domains and examine how cross-domain interactions evolve throughout training.

\paragraph{Fusion paradigms.}
Existing approaches to serving all domains with one model broadly fall into three paradigms.
\textbf{Merge} requires no further training and instead combines the experts' \textit{task vectors}---their weight differences from the base model---into a single update to the base weights~\citep{ilharco2023editing}.
Methods differ in how they weight, sparsify, and resolve conflicts among the expert updates~\citep{wortsman2022model,yadav2023ties}.
For LoRA experts~\citep{hu2022lora}, integration can be performed at the adapter level through either composition~\citep{zhang2023composing} or routing over multiple adapters~\citep{mindlab2026macaronv1preview}.
\textbf{Mix RL} instead pools data from all domains into a single training run, allowing different domains to interact throughout optimization and requiring careful data mixing~\citep{huang2026step}.
On-policy distillation (OPD) trains a student on its own rollouts under teacher supervision~\citep{lu2025onpolicydistillation, li2026rethinking}.
In a multi-teacher setting (\textbf{MOPD}), domain-specific teachers provide a natural way to consolidate specialized models~\citep{xiao2026mimo, xu2026deepseek}.
Prior work has largely studied these approaches in isolation on different backbones and benchmarks, making direct comparison difficult.
A closely related study by~\citet{wang2026mix} compares mixed multi-domain RLVR with post-hoc fusion across multiple domains in a focused setting.
In contrast, we provide a systematic comparison of Merge, Mix RL and MOPD as three distinct paradigms, analyzing their effectiveness and underlying parameter-space behavior.

\section{Preliminary}
\label{sec: preliminary}
\paragraph{Problem setup.}
Let \(\theta_0\) denote the base model parameters, and write \(\pi_\theta\) for the policy induced by parameters \(\theta\).
We consider \(N\) domains, each associated with a prompt dataset \(\mathcal{D}_i\) and a verifier \(v_i\) that scores prompt-response pairs.
We study the transition from \(N\) separately trained domain experts to a unified generalist model that performs well across all \(N\) domains.

\paragraph{Per-domain RLVR.}
Starting from \(\theta_0\), each expert is trained with RLVR on a single domain.
In group-based RLVR methods such as GRPO~\citep{shao2024deepseekmath}, the policy samples a group of \(G\) responses \(O = \{o_1, \dots, o_G\}\) for each prompt \(q \in \mathcal{D}_i\), and each response receives a reward \(r_j = v_i(q, o_j)\).
Each response is then scored relative to its group~\citep{liu2025understanding},
{
\footnotesize
\begin{equation}
    \hat{A}_j = r_j - \frac{1}{G}\sum_{k=1}^{G} r_k,
\end{equation}
}%
and this advantage is applied to each token in \(o_j\) through a policy gradient update.
We write \(\theta_i\) for the parameters of the resulting expert for domain \(i\), and \(\tau_i = \theta_i - \theta_0\) for its task vector~\citep{ilharco2023editing}, the displacement that training applied to the base weights.
Expert training may update all parameters or instead use Low-Rank Adaptation (LoRA)~\citep{hu2022lora}.
LoRA freezes \(\theta_0\) and parameterizes the task vector as \(\tau_i=\tfrac{\alpha}{r}B_iA_i\), reducing memory and optimizer-state requirements while matching full-parameter RL tuning~\citep{schulman2025lora}, making it a practical alternative for training domain experts.
We consider three fusion paradigms distinguished by what they reuse: \textbf{Merge} operates on the \(\tau_i\), \textbf{Mix RL} trains on the \(\mathcal{D}_i\), and \textbf{MOPD} draws on both.

\paragraph{Merge.}
Merge combines the \(N\) experts into a single model without further training by folding their task vectors back into the base weights~\citep{ilharco2023editing}:
{
\footnotesize
\begin{equation}
    \theta_{\mathrm{merge}} = \theta_0 + \mathcal{F}(\tau_1, \dots, \tau_N),
\end{equation}
}%
where \(\mathcal{F}\) is a combination rule over the task vectors that weights and reconciles the expert updates, and Appendix~\ref{app: merge_methods} describes the choices we compare.
For LoRA experts, \(\mathcal{F}\) can act on the reconstructed \(\tau_i\)~\citep{mangrulkar2022peft} or directly on \(A_i\) and \(B_i\)~\citep{zhang2023composing}.
Regardless of its form, \(\mathcal{F}\) produces a joint update that must represent all \(N\) experts, each of which influences the result only through its task vector \(\tau_i\).

\paragraph{Mix RL.}
Mix RL does not train separate experts.
Starting from \(\theta_0\), Mix RL trains on the union \(\bigcup_i \mathcal{D}_i\), drawing prompts from domain \(i\) with probability \(p_i\).
Each prompt is scored by its domain verifier \(v_i\), so the shared policy receives domain-specific reward signals within one optimization trajectory.
The proportions \(p_i\) therefore specify the domain composition of the shared training stream~\citep{li2025can}.
As all domains share the same weights and gradient steps, cross-domain interactions occur during training rather than after it.

\paragraph{MOPD.}
On-policy distillation trains a student under teacher supervision on states visited by its own policy.
Multi-teacher OPD (MOPD)~\citep{xiao2026mimo} uses the \(N\) experts as domain teachers, with \(\pi_{\theta_i}\) supervising the student on prompts \(q\) from the corresponding domain \(i\).
For a student rollout \(o \sim \pi_{\theta}(\cdot \mid q)\), let \(s_t = (q, o_{<t})\) denote the state at token \(t\).
MOPD minimises the reverse KL from the student to its domain teacher at every visited state~\citep{lu2025onpolicydistillation}:
{
    \footnotesize
\begin{equation}
    \mathcal{L}(\theta) = \mathbb{E}_{q,\, o \sim \pi_{\theta}} \left[ \sum_{t=1}^{|o|} D_{\mathrm{KL}}\big(\pi_{\theta}(\cdot \mid s_t) \,\|\, \pi_{\theta_i}(\cdot \mid s_t)\big) \right].
\end{equation}
}%
MOPD therefore consolidates the \(N\) specialized policies into one student while keeping supervision on-policy.
No verifier reward enters this objective, so the student learns to reproduce teacher behavior rather than directly optimize task success.

\section{Experiment}
\label{sec: experiment}
\subsection{Experimental setup}

\paragraph{Models and training datasets.}
To study multi-domain RLVR fusion across different model sizes, we use Qwen3-4B-Instruct-2507 and Qwen3-8B (non-thinking mode)~\citep{yang2025qwen3} as backbones\footnote{Although both models are instruction-tuned, Qwen3-4B-Instruct-2507 is a later iteration and, despite its smaller size, outperforms Qwen3-8B across a wide range of benchmarks.}.
To cover a broad set of capabilities, following~\citet{blakeman2025nemotron}, we construct training data spanning five domains:
\begin{inparaenum}[\it 1)]
\item \textbf{Math}, where we filter Polaris~\citep{Polaris2025} by difficulty, retaining \(38{,}131\) hard examples;
\item \textbf{Science}, where we similarly retain \(50{,}000\) hard examples from OpenScienceReasoning-2~\citep{open_science_reasoning_2_2025};
\item \textbf{Coding}, using \(19{,}169\) examples from CodeContests~\citep{li2022competition} and Open-R1~\citep{penedo2025codeforces};
\item \textbf{Instruction Following} (IF), using \(16{,}575\) examples from WildChat-1M~\citep{zhao2024wildchat} paired with instructions from Open-Instruct~\citep{lamberttulu}; and
\item \textbf{Agent}, using \(10{,}229\) examples from WorkplaceAssistant~\citep{blakeman2025nemotron}.
\end{inparaenum}
For Mix RL and MOPD, we follow~\citet{wang2026mix} and construct a mixed corpus of \(87{,}699\) examples with domain proportions of \(25\%\) math, \(22\%\) science, \(22\%\) code, \(19\%\) instruction following, and \(12\%\) agent.
Appendix~\ref{app: data_processing} provides further details on data processing.

\paragraph{Fusion setups.}
All three paradigms start from a common base model and cover the same five domains, differing only in how those domains are brought together.
For each domain, we train both a full-parameter expert and a LoRA expert with RLVR, yielding two sets of five experts.
For Merge, we follow~\citet{ilharco2023editing} and apply \textit{Task Arithmetic} separately to each set, with \(\mathcal{F} = \lambda \sum_i \tau_i\) and \(\lambda = 0.6\).
Appendix~\ref{app: merge_methods} sweeps \(\lambda\) and compares a broader set of merging methods.
Mix RL performs a single GRPO run on the mixed corpus described above.
MOPD distills the five full-parameter experts into one student.
For each student-sampled token \(o_t\), the log-probability ratio \(\log \pi_{\theta}(o_t \mid s_t) - \log \pi_{\theta_i}(o_t \mid s_t)\) estimates its reverse-KL contribution and provides a dense training signal~\citep{li2026rethinking}.
Appendix~\ref{app: implementation_details} provides further implementation details and hyperparameters.

\paragraph{Evaluation.}
To evaluate model performance across the five target domains, we consider a diverse suite of benchmarks:
\begin{inparaenum}[\it 1)]
\item \textbf{Math}: AIME 2025 and 2026~\citep{aime_2025_2026};
\item \textbf{Science}: GPQA-Diamond~\citep{rein2024gpqa};
\item \textbf{Coding}: LiveCodeBench v5 and v6~\citep{jainlivecodebench};
\item \textbf{Instruction Following}: IFEval~\citep{zhou2023instruction} and IFBench~\citep{pyatkin2026generalizing}; and
\item \textbf{Agent}: BFCL v3~\citep{patil2025bfcl}.
\end{inparaenum}
To reduce decoding variance, we sample \(32\) outputs per prompt and report \(\mathrm{mean@32}\).
At inference, we use a temperature of \(0.6\) and top-\(p\) of \(0.95\).

\subsection{Main results}
\label{subsec: main_results}
\begin{table*}[t]
    \caption{Main results for the three multi-domain RLVR fusion paradigms across five target domains. The base model and per-domain RL serve as baselines, and all scores are reported as \(\mathrm{mean@32}\). Column shading groups the eight benchmarks by domain, and \textit{Avg.} reports their mean. Boldface marks the best fusion result in each column. Each \textit{Per-domain RL} row aggregates five models rather than reporting a single model: each entry gives the performance of the expert trained for the corresponding benchmark domain. Appendix~\ref{app: expert_cross_domain} reports complete cross-domain results for all experts. \textit{Tuning} distinguishes full-parameter from LoRA experts and, for Merge, identifies the expert set being merged; Mix RL and MOPD are full-parameter throughout.}
    \label{tab: main_results}
    \begin{center}
    
    \resizebox{\linewidth}{!}{
    \begin{tabular}{lcccccccccc}
    \toprule
    & \textbf{Tuning} & \cellcolor{cmath!15}\textbf{AIME25} & \cellcolor{cmath!15}\textbf{AIME26} & \cellcolor{cscience!15}\textbf{GPQA} & \cellcolor{ccode!15}\textbf{LCBv5} & \cellcolor{ccode!15}\textbf{LCBv6} & \cellcolor{cif!15}\textbf{IFEval} & \cellcolor{cif!15}\textbf{IFBench} & \cellcolor{cagent!15}\textbf{BFCLv3} & \textbf{Avg.} \\
    \midrule
    \multicolumn{11}{c}{\textit{Qwen3-4B-Instruct-2507}} \\
    \midrule
    Base & -- & \rowavg{49.1}{53.1}{57.2}{58.9}{59.6}{83.8}{30.3}{63.9} \\
    \addlinespace[2pt]
    \multicolumn{1}{l|}{\multirow{2}{*}{+ Per-domain RL}} & Full & \rowavg{52.3}{58.4}{62.2}{63.4}{63.7}{89.8}{49.0}{72.2} \\
    \multicolumn{1}{l|}{}                                 & LoRA & \rowavg{51.5}{57.1}{61.4}{63.2}{63.2}{90.8}{50.2}{72.3} \\ 
    \addlinespace[2pt]
    \multicolumn{11}{l}{\textbf{\textit{Fusion paradigms}}} \\
    \multicolumn{1}{l|}{\multirow{2}{*}{Merge}}           & Full & \rowavg[\textbf]{\bd{54.9}}{\bd{61.6}}{\bd{63.4}}{\bd{64.7}}{63.9}{88.3}{41.0}{71.8} \\
    \multicolumn{1}{l|}{}                                 & LoRA & \rowavg{54.3}{59.6}{61.9}{63.4}{63.2}{89.0}{41.9}{68.6} \\ 
    \addlinespace[2pt]
    Mix RL & Full & \rowavg{53.0}{59.0}{61.1}{63.9}{\bd{64.1}}{85.9}{39.8}{71.3} \\
    \addlinespace[2pt]
    MOPD   & Full & \rowavg{51.2}{57.5}{62.3}{63.5}{63.1}{\bd{89.2}}{\bd{47.3}}{\bd{72.4}} \\
    \midrule
    \multicolumn{11}{c}{\textit{Qwen3-8B (non-thinking)}} \\
    \midrule
    Base & -- & \rowavg{19.0}{14.6}{46.0}{43.0}{45.9}{83.3}{25.9}{58.6} \\
    \addlinespace[2pt]
    \multicolumn{1}{l|}{\multirow{2}{*}{+ Per-domain RL}} & Full & \rowavg{39.2}{39.9}{52.5}{50.9}{50.9}{88.9}{42.8}{64.9} \\
    \multicolumn{1}{l|}{}                                 & LoRA & \rowavg{38.1}{39.6}{52.4}{50.9}{52.3}{89.0}{42.0}{63.5} \\
    \addlinespace[2pt]
    \multicolumn{11}{l}{\textbf{\textit{Fusion paradigms}}} \\
    \multicolumn{1}{l|}{\multirow{2}{*}{Merge}}           & Full & \rowavg{41.4}{43.3}{54.4}{\bd{50.9}}{\bd{51.4}}{87.6}{37.0}{63.5} \\
    \multicolumn{1}{l|}{}                                 & LoRA & \rowavg{37.1}{40.4}{54.0}{50.0}{51.0}{88.2}{37.5}{63.8} \\
    \addlinespace[2pt]
    Mix RL & Full & \rowavg[\textbf]{\bd{45.7}}{\bd{47.2}}{\bd{54.5}}{50.7}{51.1}{83.9}{37.9}{\bd{65.6}} \\
    \addlinespace[2pt]
    MOPD   & Full & \rowavg{37.1}{41.0}{52.2}{50.4}{51.1}{\bd{88.7}}{\bd{42.8}}{62.2} \\
    \bottomrule
    \end{tabular}
    }
    \end{center}
\end{table*}
    
We report the main results in Table~\ref{tab: main_results} and summarize our findings below:
\begin{inparaenum}[\it 1)]
\item
\textbf{Every trained setting improves on the base model.}
The \textit{Per-domain RL} row reports each expert on its own domain only, where every entry gains \(3.2\) to \(18.7\) points over the base model on 4B and \(5.0\) to \(25.3\) points on 8B (Appendix~\ref{app: expert_cross_domain} reports each expert on all benchmarks).
LoRA tracks full-parameter tuning within \(1.4\) points on every benchmark, so both provide comparable experts for fusion.
The fusion rows below, which fold these experts or their data into one model, likewise outperform the base model, on average by \(5.3\) to \(6.7\) points on 4B and \(10.8\) to \(12.6\) on 8B.
What separates the paradigms is how closely they match the experts, so we compare them against the \textit{Per-domain RL} row.
\item
\textbf{Merge and Mix RL redistribute the experts' gains rather than inherit them intact.}
One fuses task vectors and the other datasets, yet they produce similar patterns of gains and losses across domains.
On reasoning domains such as math and code, both paradigms either outperform the per-domain experts or stay within \(1.5\) points of them on both backbones, while their overall averages remain within \(1.6\) points of the expert average.
By contrast, they fall short on instruction following, trailing the experts by \(4.9\) to \(9.2\) points on IFBench.
The two differ in transfer strength rather than direction: Mix RL clears the math expert by \(6.5\) and \(7.3\) points on the two AIME sets at 8B, the largest margin over an expert in the table, while Merge's reasoning gains are more moderate but hold on both backbones.
Section~\ref{sec: analysis} shows that a domain's retained gain depends on its relatedness to the others, which is why the two largely agree despite operating on task vectors and datasets respectively.
\item
\textbf{MOPD matches its teachers but does not surpass them.}
MOPD consistently remains close to its teachers across all domains, avoiding the domain-level losses observed with the other two paradigms.
However, it does not outperform its teachers on average for either backbone.
This limitation follows from the standard OPD objective, as the student is optimized to match teacher behavior on its own samples rather than to surpass its teachers.
The teachers therefore provide a natural performance boundary, with no explicit learning signal that encourages the student to surpass them~\citep{yang2026learning}.
\end{inparaenum}

\section{Analysis}
\label{sec: analysis}
\begin{figure*}[t]
    \centering
    \includegraphics[width=\linewidth]{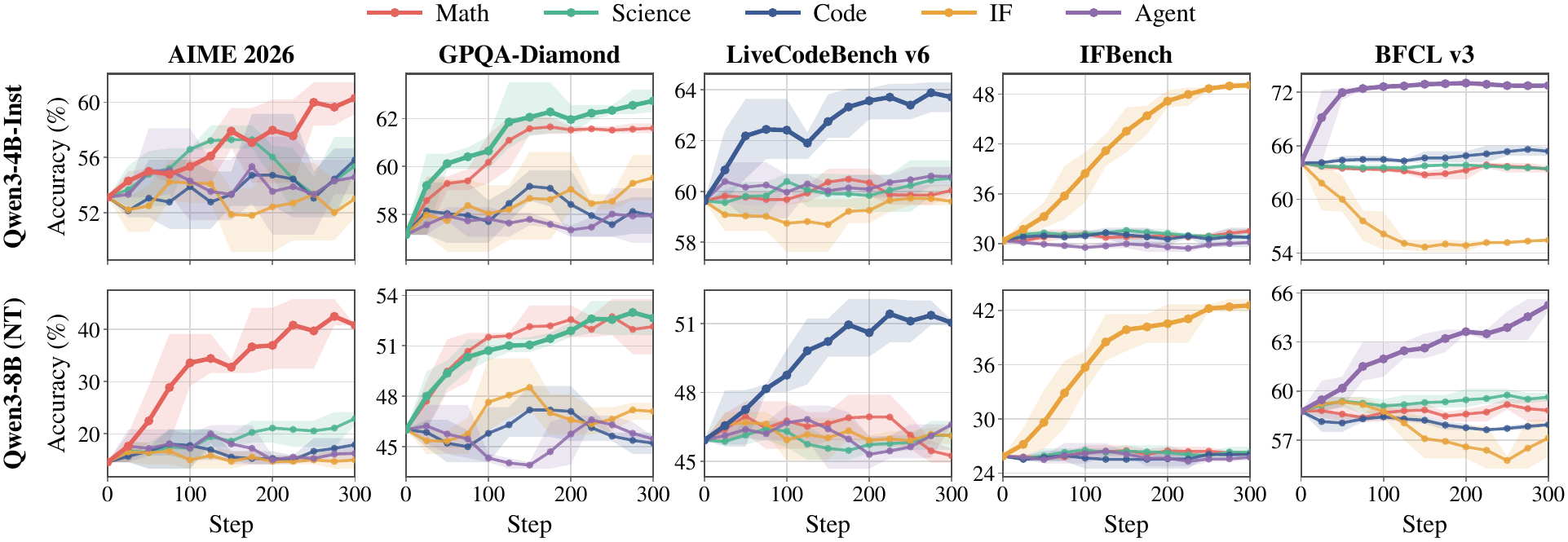}
    \vspace{-1.5em}
    \caption{Cross-domain effects of per-domain RLVR. Each panel tracks all five experts on one benchmark throughout their respective training runs. The curve for the expert trained on the corresponding domain is bolded. Curves report \(\mathrm{mean@4}\) performance.}
    \label{fig:expert_cross_domain}
\end{figure*}

\begin{figure*}[t]
    \centering
    \includegraphics[width=\linewidth]{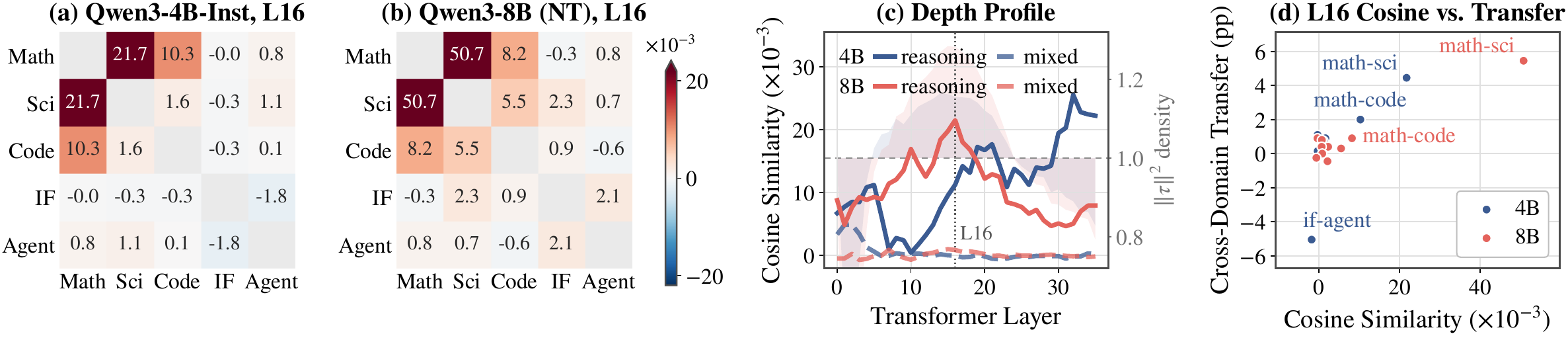}
    \vspace{-1.5em}
    \caption{Geometry of the five task vectors \(\tau_i\). \textbf{(a,b)} Pairwise cosine similarity at layer \(16\); the unit diagonal is left blank. \textbf{(c)} The same quantity across all layers, averaged over three reasoning pairs (solid) and seven pairs involving instruction following or agent (dashed). The shaded band (right axis) gives each layer's squared displacement relative to its parameter share, exceeding \(1\) where RL moves a layer more than its size predicts. \textbf{(d)} Layer-\(16\) cosine similarity versus cross-domain transfer measured in behavior, symmetrized across directions to give one point per pair and backbone.}
    \label{fig:task_vector_cosine}
\end{figure*}

\subsection{How the five domains interact}
\label{subsec: domain_interaction}

To further investigate the per-domain outcomes of the three fusion paradigms, we examine how the five domains relate from two perspectives, first in behavior and then in weight space.

\paragraph{Domain relations in behavior.}
We track each expert's cross-domain performance throughout training in Figure~\ref{fig:expert_cross_domain}, and find interactions that are neither uniform nor symmetric.
As shown, math and science help each other on both backbones: training on math raises GPQA by about \(5\) points, close to the gain achieved by training on science itself, and training on science also raises AIME 2026 performance.
The clearest negative transfer is from instruction following to agent use.
On the 4B backbone, the IF expert lowers BFCL v3 performance by \(9.8\) points, and this loss deepens as training proceeds.
These effects divide the five domains according to what they require of the model.
Math, science and code are reasoning-intensive, and transfer among them is largely positive.
Instruction following and agent use depend more on task-specific demands than on intensive reasoning.
The former requires the model to parse and honour constraints, whereas the latter requires it to model an environment and plan interactions~\citep{wang2026mix}.
Their effects on the other three stay within \(2.0\) points on both backbones, and neither helps the other.

\paragraph{Domain relations in weight space.}
The behavioural view shows which domains help one another, but not whether that structure is already present in the parameters that fusion combines.
Since merging operates on the task vectors \(\tau_i\) directly, we measure their geometry.
Let \(\tau_i^{S}\) denote the restriction of \(\tau_i\) to a parameter set \(S\).
We compute \(\cos\nolimits_{S}(i, j) = \frac{\langle \tau_i^{S},\, \tau_j^{S} \rangle}{\|\tau_i^{S}\|\,\|\tau_j^{S}\|}\) over its coordinates, taking \(S\) to be a single transformer layer.
To select the layer, we compare relative RLVR displacement across layers.
For each layer \(\ell\), we divide its share of \(\|\tau_i\|^2\) by its share of model parameters and average the ratio across the five experts.
Values above \(1\) indicate greater displacement than expected from layer size.
This quantity peaks in the middle of both networks (Figure~\ref{fig:task_vector_cosine}(c), shaded), so we report layer \(16\), which~\citet{zhang2026one} identify as one of the highest-contribution layers.

The same grouping appears in weight space.
In units of \(10^{-3}\), the layer-\(16\) cosine similarity reaches \(21.7\) on 4B and \(50.7\) on 8B for math--science, and \(10.3\) and \(8.2\) for math--code, while every pair involving instruction following or agent has magnitude at most \(2.3\) (Figure~\ref{fig:task_vector_cosine}(a,b)).
Figure~\ref{fig:task_vector_cosine}(c) repeats the measurement for every layer, averaging the three reasoning-only pairs against the seven involving instruction following or agent (mixed), and the two groups stay apart across the whole network, so the separation is not an artifact of a specific layer.
Figure~\ref{fig:task_vector_cosine}(d) plots each pair's layer-\(16\) cosine similarity against its cross-domain transfer, symmetrized over the two directions: math-science sits at the top right on both backbones and IF-agent at the bottom left on 4B, so pairs that move together in weight space also help each other in behavior.
Instruction following therefore has the update direction least aligned with those of other domains, and this structure is visible in the weights before fusion.
The same geometry also determines what Merge retains.
Summing the five task vectors preserves only part of each expert's displacement, with near-orthogonality keeping the retained share similar across domains and backbones.
How much of an expert's \textit{gain} survives therefore varies by domain.
Merge matches or exceeds reasoning-domain experts on both backbones because each reasoning domain benefits from the other two updates.
By contrast, instruction following shares little with those updates and retains roughly \(60\%\) of its IFBench gain.

\begin{finding}
Cross-domain transfer is structured: reasoning-intensive domains tend to reinforce one another, while \textit{IF} and \textit{Agent} remain orthogonal in both behavior and task-vector geometry.
\end{finding}

\begin{figure*}[t]
    \centering
    \includegraphics[width=\linewidth]{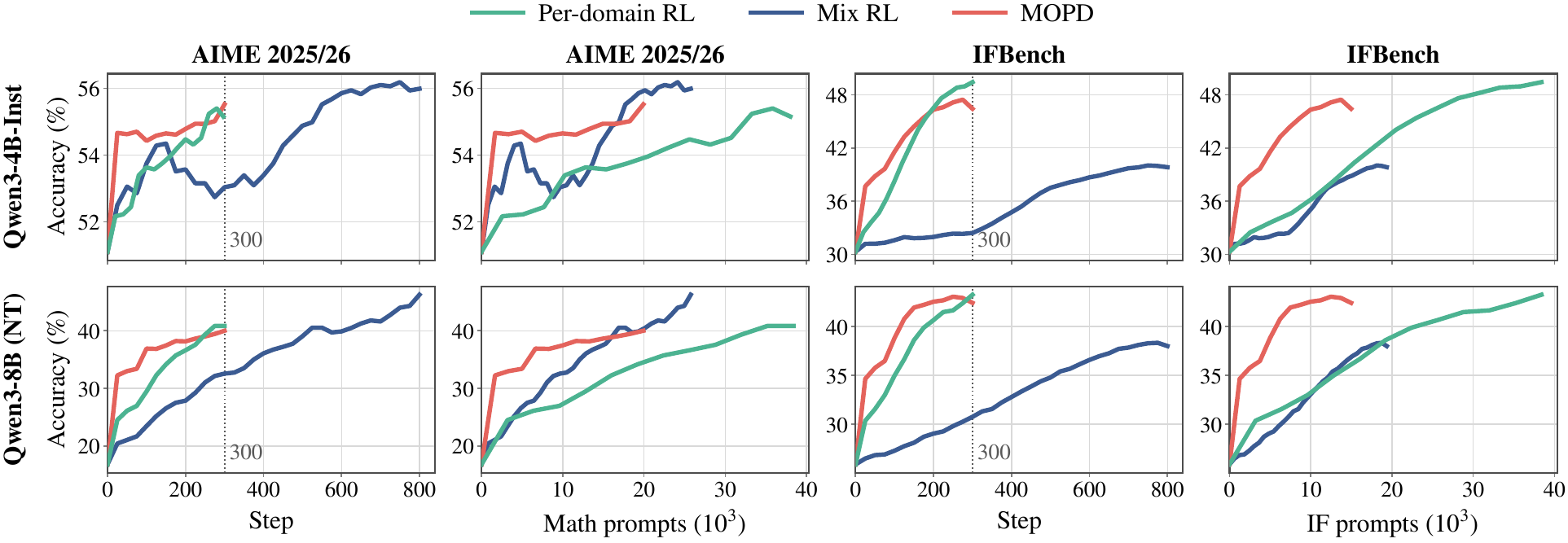}
    \caption{Training performance of per-domain RL, Mix RL and MOPD. Merge is omitted because it has no training trajectory. Within each row, the first pair of panels reports AIME 2025/26 and the second reports IFBench. The two panels in each pair plot the same curves against optimization steps and the number of prompts sampled from the corresponding domain.}
    \label{fig:fusion_speed}
    \vspace{-1em}
\end{figure*}

\begin{table}[t]
    \caption{Requirements and costs of the three paradigms. \textit{Experts} indicates whether the paradigm requires the five per-domain models; \textit{Supervision} denotes the learning signal; and \textit{Deploy} indicates the number of models served at inference. \textit{Fusion} reports the cost of the fusion stage alone, while \textit{Total} additionally includes the cost of training the five experts when required. The final two columns report \textit{Total} relative to per-domain RL. Appendix~\ref{app: implementation_details} reports individual expert costs in detail.}
    \label{tab: paradigm_cost}
    \begin{center}

    \resizebox{\linewidth}{!}{
    \begin{tabular}{lccccccccc}
    \toprule
    & \multicolumn{3}{c}{\textbf{Requirements}} & \multicolumn{2}{c}{\textbf{Fusion GPU-h}} & \multicolumn{2}{c}{\textbf{Total GPU-h}} & \multicolumn{2}{c}{\textbf{Total} (\(\times\))} \\
    \cmidrule(lr){2-4} \cmidrule(lr){5-6} \cmidrule(lr){7-8} \cmidrule(lr){9-10}
    & \textbf{Experts} & \textbf{Supervision} & \textbf{Deploy} & 4B & 8B & 4B & 8B & 4B & 8B \\
    \midrule
    Per-domain RL & --           & Verifiable reward & 5 & --     & --     & 5{,}220 & 4{,}670 & \(1.00\) & \(1.00\) \\
    \addlinespace[2pt]
    \multicolumn{10}{l}{\textbf{\textit{Fusion paradigms}}} \\
    Merge         & Required     & None              & 1 & \(\approx 0\) & \(\approx 0\) & 5{,}220 & 4{,}670 & \(1.00\) & \(1.00\) \\
    \addlinespace[2pt]
    Mix RL        & Not required & Verifiable reward & 1 & 3{,}042 & 3{,}138 & 3{,}042 & 3{,}138 & \(0.58\) & \(0.67\) \\
    \addlinespace[2pt]
    MOPD          & Required     & Teacher logprobs  & 1 & \phantom{0{,}}741 & \phantom{0{,}}899 & 5{,}960 & 5{,}569 & \(1.14\) & \(1.19\) \\
    \bottomrule
    \end{tabular}
    }

    \end{center}
\end{table}

\subsection{How the three paradigms compare}
\label{subsec: fusion_speed}

The previous subsection focused on relations among the five domains; we now ask how the three paradigms convert training resources into per-domain performance and what each requires.
We focus on math and IF, which our analysis finds representative of contrasting interaction patterns.

\paragraph{Convergence follows the domain relations.}
Figure~\ref{fig:fusion_speed} tracks per-domain RL, Mix RL and MOPD throughout training against both optimization steps and the number of domain-specific prompts.
We include the second axis because one optimization step processes different amounts of domain-specific data across methods: an expert devotes its entire batch to its domain, whereas a mixed run allocates only part of each batch to any one domain.
Against steps, Mix RL converges most slowly in every panel.
At step \(300\), when the other two methods end their runs, it trails both but closes the gap with further training.
Against prompts, the curves reflect the domain relations identified in Section~\ref{subsec: domain_interaction}.
The update direction for instruction following is weakly aligned with those of the other four domains, so progress depends primarily on the number of instruction-following prompts seen.
Mix RL therefore closely tracks the expert on both backbones.
The IFBench shortfall, \(9.2\) points at 4B and \(4.9\) at 8B, reflects both instruction following's \(19\%\) share of the mixed corpus and the absence of positive transfer from the other domains.
Math differs because the other domains contribute positively.
At 8B, Mix RL outperforms the expert on the AIME benchmarks while drawing only \(25.8\)k math prompts, compared with the expert's \(38.4\)k.
Mix RL therefore requires careful data mixing, as domain proportions directly shape per-domain outcomes under a fixed training budget.

\paragraph{MOPD converges fastest but remains bounded by its teachers.}
MOPD stays above per-domain RL at matched data exposure in every panel and converges early on both axes: at least \(70\%\) of its total gain has already been achieved at step \(100\), one third of the way through training.
Its gains then flatten while per-domain RL continues to improve, reflecting the boundary in Section~\ref{subsec: main_results}---matching teacher behavior on student-generated samples lets MOPD reach teacher performance with less data but not surpass it.
MOPD splits its batch across the five domains as Mix RL does, but each prompt carries dense teacher supervision.
With a target provided at every visited state, the student follows a known direction rather than discovering one through exploration, allowing it to make more progress per step with fewer domain-specific prompts than the expert.
The same experts can also be reused without further training by merging their task vectors in parameter space rather than distilling their behavior on student-generated samples.

\paragraph{What the three paradigms ask for.}
Table~\ref{tab: paradigm_cost} summarizes the requirements and costs of the three paradigms.
As a reference, per-domain RL costs \(5{,}220\) and \(4{,}670\) GPU-hours for 4B and 8B, respectively, while leaving five models to serve.
Once these experts exist, Merge returns a single model with only minutes of additional arithmetic, without further optimization or access to original training data.
By contrast, Mix RL is the only paradigm that efficiently produces a single model without first training experts, costing \(0.58\times\) and \(0.67\times\) the reference on 4B and 8B.
Although MOPD's fusion stage costs under \(0.2\times\) the reference on both backbones, training its five teachers raises end-to-end costs to \(1.14\times\) and \(1.19\times\), the highest of the three paradigms.
We report costs for full-parameter runs, as LoRA primarily changes memory use rather than training time because rollout generation dominates each step.
The paradigms also differ in supervision required at fusion and its source.
Mix RL uses the same domain-specific verifiers as per-domain RL throughout joint training, whereas neither expert-based paradigm uses verifier rewards during fusion.
Merge operates on weights alone, while MOPD matches teacher behavior on student-generated samples.

\begin{finding}
Convergence reflects domain relations, while the paradigms differ in how they trade resources for performance: Mix RL avoids experts but requires more steps and careful data mixing, MOPD converges fastest but is teacher-bounded, and Merge is nearly free once experts exist.
\end{finding}

\begin{figure*}[t]
    \centering
    \includegraphics[width=\linewidth]{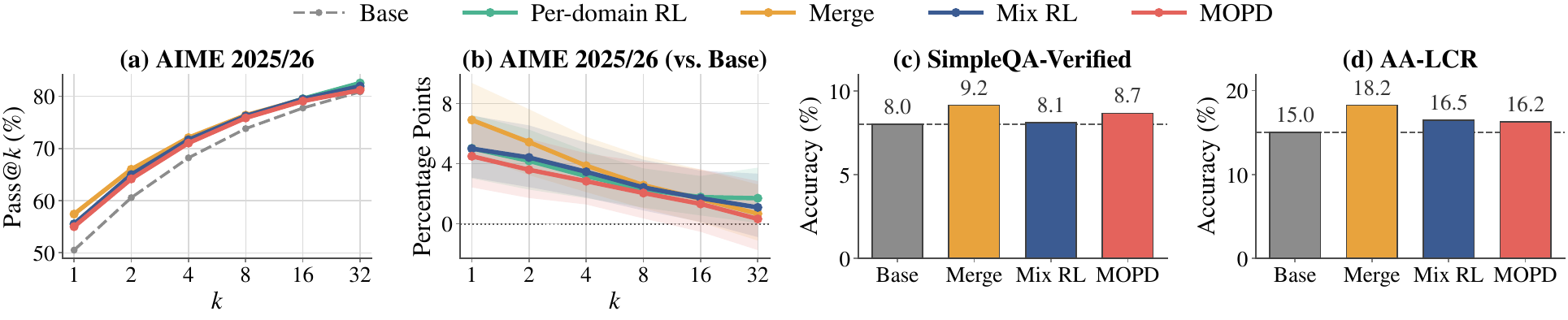}
    \vspace{-1.5em}
    \caption{Solution coverage in math and capability retention on the 4B backbone. \textbf{(a,b)} \(\mathrm{pass@}k\) on AIME 2025/26, shown as absolute performance and paired differences from the base model. \textbf{(c,d)} Held-out SimpleQA-Verified and AA-LCR performance, with dashed lines marking the base model.}
    \label{fig:coverage_and_retention}
    \vspace{-1em}
\end{figure*}

\subsection{What fusion changes and leaves alone}
\label{subsec: coverage_retention}

The analysis so far has focused on the five target domains, leaving two questions: whether fusion expands the set of solvable problems or only increases the probability of finding existing solutions, and whether it degrades capabilities outside these domains.
To answer both questions, we use the 4B backbone to measure \(\mathrm{pass@}k\) on math and accuracy on two capabilities held out from training.

\paragraph{Fusion reweights solutions within the target domains.}
For the first question, we track \(\mathrm{pass@}k\), which counts a problem as solved when any of \(k\) samples is correct~\citep{chen2021evaluating}, to distinguish reweighting from expanded solution coverage.
Figure~\ref{fig:coverage_and_retention}(a,b) reports this metric on AIME 2025/26, using outputs sampled at temperature \(1.0\) to better expose solution coverage.
At \(k = 1\), all four trained settings improve on the base model by \(4.5\) to \(6.9\) points, while the three fusion paradigms differ by \(2.4\) points with overlapping intervals.
The fusion paradigms' advantage then decays monotonically with \(k\), and by \(k = 32\) none of the three remains distinguishable from the base model.
Sampling the base model a few more times therefore recovers what fusion delivers with one sample.
MOPD is especially informative given recent evidence that distillation can expand the boundary by importing patterns a teacher has and a student lacks~\citep{yue2025does}.
Such expansion does not occur here because MOPD's teachers are RLVR experts derived from the same base model and provide no new solution support to import.
Taken together, these results show that fusion primarily reweights solutions already accessible to the base model rather than adding new ones.

\paragraph{Fusion preserves capabilities outside the target domains.}
To test whether the gains come at a cost outside the five target domains, we evaluate two capabilities excluded from all training runs.
Specifically, SimpleQA-Verified~\citep{haas2025simpleqa} measures parametric factual recall using fact-seeking questions, while AA-LCR~\citep{artificialanalysis2025lcr} evaluates long-context reasoning over document sets of roughly \(100\)k tokens.
We use Qwen3-32B~\citep{yang2025qwen3} as the LLM judge for both benchmarks and report \(\mathrm{mean@4}\) in Figure~\ref{fig:coverage_and_retention}(c,d).
Appendix~\ref{app: held_out_eval} details the evaluation protocol and judge prompts.
The results show that no paradigm scores below the base model on either benchmark.
This suggests that gains on the five target domains sacrifice neither the backbone's factual knowledge nor its long-context reasoning.
This aligns with prior evidence that RLVR can improve target performance without noticeable degradation on non-target tasks~\citep{chen2026retaining}.

\begin{finding}
Fusion improves single-sample accuracy by reweighting solutions already accessible to the base model, without measurable expansion in coverage or degradation of held-out capabilities.
\end{finding}

\section{Conclusion}
\label{sec: conclusion}
We compare Merge, Mix RL and MOPD for consolidating domain-specific RLVR gains into one model.
Their outcomes reflect cross-domain relations in behavior and task-vector geometry.
Merge is nearly free with experts, Mix RL avoids expert training but depends on data allocation, and MOPD converges fastest but is teacher-bounded and costliest overall.
All three reweight existing solutions without measurable coverage gains or held-out capability losses.
Accordingly, Merge suits existing experts and tight budgets, Mix RL suits training without experts when data allocation can be tuned, and MOPD suits preserving domain gains despite its teacher ceiling and higher cost.

\bibliography{iclr2027_conference}
\bibliographystyle{iclr2027_conference}

\newpage
\appendix
\section*{Appendix}
\section{Details of Data Processing}
\label{app: data_processing}
\paragraph{Per-domain expert training data.}
In our prior experiments, we observed that models learn little from overly easy problems, since such problems provide limited learning signals~\citep{Polaris2025}.
For the \textbf{math} training set, we start from Polaris~\citep{Polaris2025}, in which each problem carries a difficulty label \(k/8\): the pass rate estimated by sampling \(8\) solutions from Deepseek-R1-Distill-Qwen-7B~\citep{guo2025deepseek}, where \(k\) is the number of correct solutions.
Because hard problems provide stronger learning signals for RL, we remove those with a pass rate \(> 4/8\), retaining \(38{,}131\) training examples.
For the \textbf{science} domain, we start from OpenScienceReasoning-2~\citep{open_science_reasoning_2_2025} and obtain difficulty labels analogously by sampling \(8\) solutions per problem from Qwen3-4B-Instruct-2507~\citep{yang2025qwen3}.
We then discard the easiest problems with a pass rate \(> 6/8\), and randomly subsample \(50{,}000\) examples for training.
For the \textbf{code}, \textbf{instruction following} and \textbf{agent} domains, we directly use the datasets provided by their original authors, comprising \(19{,}169\), \(16{,}575\), and \(10{,}229\) training examples, respectively.

\paragraph{Mix RL and MOPD training data.}
To build the mixed-training baseline, we blend the five per-domain training sets, which are identical to those used to train the single-domain experts, into a single corpus for one joint RL run.
Following the mixing ratio of~\citet{wang2026mix}, we target the proportions of \(25\%\) math, \(22\%\) science, \(22\%\) code, \(19\%\) instruction following, and \(12\%\) agent, yielding \(87{,}699\) examples in total.

\section{Implementation Details}
\label{app: implementation_details}
We implement all methods on top of the open-source framework verl~\citep{sheng2025hybridflow} and summarize the detailed hyperparameters in Table~\ref{tab: implementation_details}.
For LoRA-based RL, every expert adapter has rank \(r = 32\) and \(\alpha = 64\) and is applied to all linear layers of the backbone.
We further adopt a larger learning rate of \(2\times 10^{-5}\)~\citep{schulman2025lora}, except for the math domain, where we lower it to \(1\times 10^{-5}\) to preserve training stability.
All experiments are conducted on 32 NVIDIA H20 GPUs.

\paragraph{Training cost.}
Table~\ref{tab: paradigm_cost_detail} reports the cost of every run underlying Table~\ref{tab: paradigm_cost}.

\begin{table*}[ht]
\caption{Implementation details and training hyperparameters for per-domain RL, Mix RL and MOPD. Merge is omitted because it requires no training.}
\label{tab: implementation_details}
\begin{center}

\begin{tabular}{lccc}
\toprule
\textbf{Hyperparameter} & \textbf{Per-domain RL} & \textbf{Mix RL} & \textbf{MOPD} \\
\midrule
Algorithm              &  GRPO &  GRPO & Sampled-Token OPD\\
Rollout batch size     &  \(128\) &  \(128\) &  \(264\) \\
Mini batch size        &  \(128\) &  \(128\) &  \(264\) \\
Rollout $n$            &  \(16\) &  \(16\) &  \(4\) \\
Maximum prompt length  &  \(5{,}120\) &  \(5{,}120\) &  \(5{,}120\) \\
Maximum response length&  \(16{,}384\) &  \(16{,}384\) &  \(16{,}384\) \\
Temperature            &  \(1.0\) &  \(1.0\) &  \(1.0\) \\
Learning rate          &  \(1\times 10^{-6}\) &  \(1\times 10^{-6}\) &  \(1\times 10^{-6}\) \\
Training steps          &  \(300\) &  \(800\) &  \(300\) \\
\bottomrule
\end{tabular}
\end{center}
\end{table*}

\begin{table}[t]
    \caption{Per-run training costs underlying the aggregate costs in Table~\ref{tab: paradigm_cost}. Costs are measured through step \(300\) for the five experts and MOPD, and through step \(800\) for Mix RL.}
    \label{tab: paradigm_cost_detail}
    \begin{center}

    \begin{tabular}{lcc}
    \toprule
    & \textbf{4B GPU-h} & \textbf{8B GPU-h} \\
    \midrule
    \multicolumn{3}{l}{\textit{Per-domain RL}} \\
    \quad Math                  & 1{,}824 & 1{,}847 \\
    \quad Science               & \phantom{0{,}}849 & \phantom{0{,}}523 \\
    \quad Code                  & 1{,}435 & \phantom{0{,}}670 \\
    \quad Instruction following & \phantom{0{,}}412 & \phantom{0{,}}633 \\
    \quad Agent                 & \phantom{0{,}}700 & \phantom{0{,}}997 \\
    \quad All five              & 5{,}220 & 4{,}670 \\
    \addlinespace[2pt]
    Mix RL                      & 3{,}042 & 3{,}138 \\
    \addlinespace[2pt]
    MOPD                        & \phantom{0{,}}741 & \phantom{0{,}}899 \\
    \bottomrule
    \end{tabular}

    \end{center}
\end{table}

\section{Per-domain Experts across All Domains}
\label{app: expert_cross_domain}

\begin{figure*}[t]
    \centering
    \includegraphics[width=\linewidth]{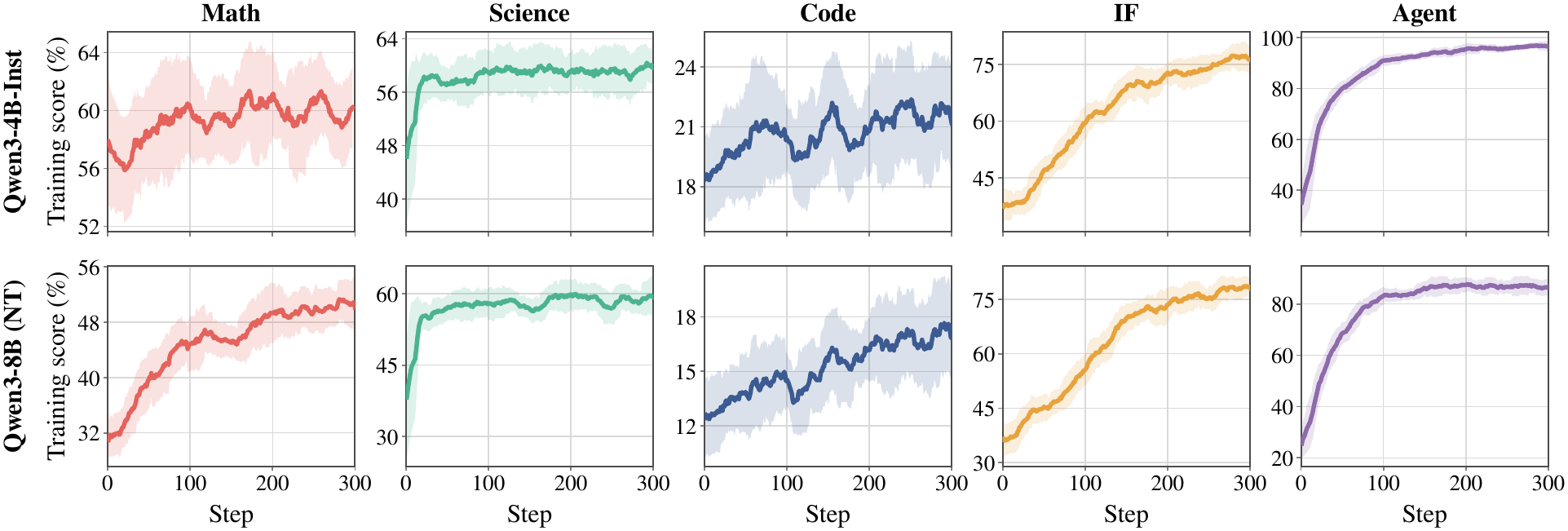}
    \caption{Training score of each per-domain expert on its own domain. Curves are smoothed by a moving average, with the band indicating standard deviation over the same window.}
    \label{fig:expert_train_score}
\end{figure*}

Table~\ref{tab: expert_cross_domain} reports each expert across the full benchmark suite, extending the \textit{Per-domain RL} row of Table~\ref{tab: main_results}, which reports each expert only on its own domain.
Figure~\ref{fig:expert_train_score} shows the training score of each run, confirming that every expert improves on the domain it was trained on.

\begin{table*}[t]
\caption{Performance of each per-domain expert across all evaluation benchmarks.}
\label{tab: expert_cross_domain}
\begin{center}

\resizebox{\linewidth}{!}{
\begin{tabular}{lcccccccccc}
\toprule
\textbf{Expert} & \textbf{Tuning} & \cellcolor{cmath!15}\textbf{AIME25} & \cellcolor{cmath!15}\textbf{AIME26} & \cellcolor{cscience!15}\textbf{GPQA} & \cellcolor{ccode!15}\textbf{LCBv5} & \cellcolor{ccode!15}\textbf{LCBv6} & \cellcolor{cif!15}\textbf{IFEval} & \cellcolor{cif!15}\textbf{IFBench} & \cellcolor{cagent!15}\textbf{BFCLv3} & \textbf{Avg.} \\
\midrule
\multicolumn{11}{c}{\textit{Qwen3-4B-Instruct-2507}} \\
\midrule
Base & -- & \rowavg{49.1}{53.1}{57.2}{58.9}{59.6}{83.8}{30.3}{63.9} \\
\addlinespace[2pt]
\cellcolor{cmath!15}                             & Full & \rowavg{52.3}{58.4}{61.8}{59.6}{60.6}{84.1}{30.8}{63.1} \\
\cellcolor{cmath!15}\multirow{-2}{*}{Math}       & LoRA & \rowavg{51.5}{57.1}{59.9}{59.6}{60.0}{83.9}{30.9}{62.6} \\
\addlinespace[2pt]
\cellcolor{cscience!15}                          & Full & \rowavg{50.4}{57.4}{62.2}{59.6}{60.9}{83.9}{30.8}{64.2} \\
\cellcolor{cscience!15}\multirow{-2}{*}{Science} & LoRA & \rowavg{47.0}{54.7}{61.4}{59.4}{59.7}{83.7}{30.5}{63.4} \\
\addlinespace[2pt]
\cellcolor{ccode!15}                             & Full & \rowavg{47.2}{56.1}{57.7}{63.4}{63.7}{83.9}{30.7}{64.9} \\
\cellcolor{ccode!15}\multirow{-2}{*}{Code}       & LoRA & \rowavg{50.7}{54.6}{57.6}{63.2}{63.2}{83.8}{31.0}{65.4} \\
\addlinespace[2pt]
\cellcolor{cif!15}                               & Full & \rowavg{48.9}{52.5}{58.9}{58.6}{59.5}{89.8}{49.0}{54.1} \\
\cellcolor{cif!15}\multirow{-2}{*}{IF}           & LoRA & \rowavg{45.6}{51.5}{59.1}{58.2}{59.7}{90.8}{50.2}{52.5} \\
\addlinespace[2pt]
\cellcolor{cagent!15}                            & Full & \rowavg{47.4}{54.5}{58.4}{59.5}{60.2}{84.0}{30.0}{72.2} \\
\cellcolor{cagent!15}\multirow{-2}{*}{Agent}     & LoRA & \rowavg{50.9}{56.5}{58.3}{61.1}{61.3}{83.6}{29.5}{72.3} \\
\midrule
\multicolumn{11}{c}{\textit{Qwen3-8B (non-thinking)}} \\
\midrule
Base & -- & \rowavg{19.0}{14.6}{46.0}{43.0}{45.9}{83.3}{25.9}{58.6} \\
\addlinespace[2pt]
\cellcolor{cmath!15}                             & Full & \rowavg{39.2}{39.9}{52.4}{43.0}{46.1}{83.0}{26.5}{58.2} \\
\cellcolor{cmath!15}\multirow{-2}{*}{Math}       & LoRA & \rowavg{38.1}{39.6}{52.6}{43.8}{45.9}{83.1}{26.2}{58.3} \\
\addlinespace[2pt]
\cellcolor{cscience!15}                          & Full & \rowavg{21.2}{19.1}{52.5}{44.2}{46.2}{83.3}{26.5}{58.8} \\
\cellcolor{cscience!15}\multirow{-2}{*}{Science} & LoRA & \rowavg{23.9}{22.7}{52.4}{44.1}{46.6}{83.4}{26.4}{58.7} \\
\addlinespace[2pt]
\cellcolor{ccode!15}                             & Full & \rowavg{18.8}{16.2}{46.3}{50.9}{50.9}{83.1}{25.7}{57.8} \\
\cellcolor{ccode!15}\multirow{-2}{*}{Code}       & LoRA & \rowavg{19.6}{16.6}{45.8}{50.9}{52.3}{83.0}{25.7}{56.8} \\
\addlinespace[2pt]
\cellcolor{cif!15}                               & Full & \rowavg{18.5}{15.8}{46.2}{42.7}{46.1}{88.9}{42.8}{57.7} \\
\cellcolor{cif!15}\multirow{-2}{*}{IF}           & LoRA & \rowavg{17.2}{13.5}{45.4}{43.0}{45.4}{89.0}{42.0}{56.9} \\
\addlinespace[2pt]
\cellcolor{cagent!15}                            & Full & \rowavg{19.9}{16.6}{46.6}{43.2}{46.2}{83.4}{25.9}{64.9} \\
\cellcolor{cagent!15}\multirow{-2}{*}{Agent}     & LoRA & \rowavg{20.3}{15.8}{46.0}{42.9}{45.8}{82.9}{25.8}{63.5} \\
\bottomrule
\end{tabular}
}
\end{center}
\end{table*}

\section{Held-out Evaluation Protocol}
\label{app: held_out_eval}

The two held-out benchmarks in Section~\ref{subsec: coverage_retention}, SimpleQA-Verified and AA-LCR, are graded by a model rather than by a rule.
We sample \(4\) responses per question and grade each with Qwen3-32B~\citep{yang2025qwen3} in non-thinking mode at temperature \(0\), using grader prompts based on those published with each benchmark~\citep{haas2025simpleqa, artificialanalysis2025lcr}.
We verify these verdicts by hand on random samples and observe no systematic grading errors.
Prompts~\ref{prompt: simpleqa} and~\ref{prompt: aa_lcr} provide the full grader prompts used for the two benchmarks.

\begin{prompt}{SimpleQA-Verified}
\label{prompt: simpleqa}
\ttfamily\small\raggedright
Your job is to look at a question, a gold target, and a predicted answer, and then assign a grade of either ["CORRECT", "INCORRECT", "NOT\_ATTEMPTED"]. First, I will give examples of each grade, and then you will grade a new example.

\medskip
[\textit{Worked examples of CORRECT, INCORRECT and NOT\_ATTEMPTED predicted answers, followed by the notes on numeric ranges, on gold targets carrying more information than the question, and on typos in names, reproduced verbatim from Appendix A of~\citealp{haas2025simpleqa}.}]

\medskip
Here is a new example. Simply reply with either CORRECT, INCORRECT, NOT ATTEMPTED. Don't apologize or correct yourself if there was a mistake; we are just trying to grade the answer.

Question: \{question\} \\
Gold target: \{target\} \\
Predicted answer: \{predicted\_answer\}

\medskip
Two of the rules above matter most for a predicted answer as long as this one may be, so note them again now that it has been read. Grade only the single candidate the answer finally commits to: it may reason at length, revise itself, and discard candidates on the way, and neither the discarded candidates nor the reasoning around them is what is being graded. And an answer denying the question's premise -- that no such person, thing or event exists, or that the question rests on a mistake -- while the gold target names one is a factual statement contradicting the gold target, so it is INCORRECT rather than NOT\_ATTEMPTED.

\medskip
Grade the predicted answer of this new question as one of: \\
A: CORRECT \\
B: INCORRECT \\
C: NOT\_ATTEMPTED

Just return the letters "A", "B", or "C", with no text around it.
\end{prompt}

\begin{prompt}{AA-LCR}
\label{prompt: aa_lcr}
\ttfamily\small\raggedright
Assess whether the following CANDIDATE ANSWER is CORRECT or INCORRECT. \\
For the CANDIDATE ANSWER to be correct, it must be consistent with the OFFICIAL ANSWER. \\
A CANDIDATE ANSWER that declines to answer, or that concludes the answer cannot be determined from the documents, is INCORRECT regardless of how well it argues the point.

\medskip
The question, for reference only: \{question\} \\
The OFFICIAL ANSWER: \{official\_answer\} \\
CANDIDATE ANSWER TO ASSESS: \{candidate\_answer\}

\medskip
Reply only with CORRECT or INCORRECT.
\end{prompt}

\section{Comparison of Merging Methods}
\label{app: merge_methods}
Table~\ref{tab: merge_methods} reports every method described below on both backbones, extending the Task-Arithmetic \textit{Merge} rows in Table~\ref{tab: main_results}.

\paragraph{Full-parameter merging.}
We merge the full-parameter experts with the MergeLM implementation~\citep{yu2024language}.
\textbf{Average}~\citep{wortsman2022model} takes the mean of the task vectors, \(\mathcal{F} = \frac{1}{N}\sum_{i} \tau_i\).
\textbf{Task Arithmetic (TA)}~\citep{ilharco2023editing} takes their sum under a shared scaling coefficient, \(\mathcal{F} = \lambda \sum_{i} \tau_i\).
\textbf{TIES}~\citep{yadav2023ties} trims each \(\tau_i\) to the \(20\%\) of entries with the largest magnitude, computed over all parameters jointly, elects one sign per parameter from the trimmed values summed across experts, and averages only the entries agreeing with it.
\textbf{DARE-TA}~\citep{yu2024language} drops each entry of \(\tau_i\) independently with probability \(p = 0.2\) and rescales the survivors by \(1/(1-p)\), leaving the update unchanged in expectation, before merging with TA.
\textbf{SCE}~\citep{wan2025fusechat} weights each \(\tau_i\) per parameter matrix by its mean squared magnitude, erases the entries whose sign disagrees with the majority sign, and normalizes by the surviving weights, keeping all positions rather than only the highest-variance ones.

\paragraph{LoRA merging.}
For LoRA experts, each task vector factorizes as \(\tau_i = B_i A_i\) with rank \(r = 32\), so the adapters themselves can be merged.
We do so with the adapter combination utilities of PEFT~\citep{mangrulkar2022peft}.
\textbf{Concat} stacks \(\{A_i\}\) and \(\{B_i\}\) along the rank dimension into a single adapter of rank \(Nr = 160\), reproducing the weighted sum of the updates exactly.
\textbf{SVD} instead forms \(\sum_i w_i B_i A_i\) explicitly and truncates it back to rank \(r\).
The remaining methods apply the weighting and sparsification rules of their full-parameter counterparts to \(A_i\) and \(B_i\) directly.

\begin{wrapfigure}{r}{0.4\linewidth}
    \vspace{-\intextsep}
    \centering
    \includegraphics[width=\linewidth]{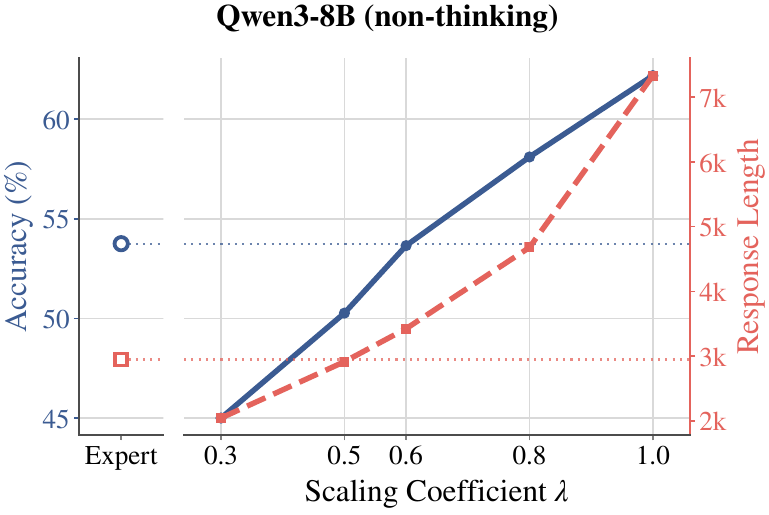}
    \caption{Accuracy and response length of the Qwen3-8B TA merge versus the scaling coefficient \(\lambda\), averaged over the evaluation sets across domains.}
    \label{fig:lambda_sweep}
\end{wrapfigure}

\paragraph{Scaling coefficient \(\lambda\).}
Figure~\ref{fig:lambda_sweep} sweeps \(\lambda\) for TA on Qwen3-8B.
Accuracy improves with \(\lambda\), but response length grows faster, from \(2{,}042\) tokens at \(\lambda = 0.3\) to \(7{,}334\) at \(\lambda = 1.0\), passing the per-domain experts (hollow markers) already at \(\lambda \approx 0.5\).
As Qwen3-8B is a hybrid-thinking model~\citep{yang2025qwen3}, we assume that the extra tokens partly reflect leakage into thinking mode, as a large \(\lambda\) scales up the merge update until it washes out the pattern that holds the model in non-thinking mode.
To verify this, we count responses that emit their own reasoning trace despite the non-thinking template.
Their frequency is negligible for small \(\lambda\) but rises markedly as \(\lambda\) grows (from \({\sim}0\%\) to \({\sim}66\%\)).
To prevent thinking-mode leakage from confounding the comparison, we report \(\lambda = 0.6\) in our main experiments.
\begin{table*}[t]
    \caption{Comparison of merging methods. Within each group, methods combine the same five expert task vectors. Concat and SVD apply only to LoRA experts. TA and DARE-TA use \(\lambda = 0.6\). For each backbone, the two \textit{+ Per-domain RL} rows report scores for the full-parameter and LoRA experts on their respective domains. Methods marked with \textsuperscript{\dag} are reported as \textit{Merge} in Table~\ref{tab: main_results}.}
    \label{tab: merge_methods}
    \begin{center}
    
    \resizebox{\linewidth}{!}{
    \begin{tabular}{lccccccccc}
    \toprule
    \textbf{Method} & \cellcolor{cmath!15}\textbf{AIME25} & \cellcolor{cmath!15}\textbf{AIME26} & \cellcolor{cscience!15}\textbf{GPQA} & \cellcolor{ccode!15}\textbf{LCBv5} & \cellcolor{ccode!15}\textbf{LCBv6} & \cellcolor{cif!15}\textbf{IFEval} & \cellcolor{cif!15}\textbf{IFBench} & \cellcolor{cagent!15}\textbf{BFCLv3} & \textbf{Avg.} \\
    \midrule
    \multicolumn{10}{c}{\textit{Qwen3-4B-Instruct-2507}} \\
    \midrule
Base & \rowavg{49.1}{53.1}{57.2}{58.9}{59.6}{83.8}{30.3}{63.9} \\
+ Per-domain RL (Full)                   & \rowavg{52.3}{58.4}{62.2}{63.4}{63.7}{89.8}{49.0}{72.2} \\
    + Per-domain RL (LoRA)                   & \rowavg{51.5}{57.1}{61.4}{63.2}{63.2}{90.8}{50.2}{72.3} \\
    \addlinespace[2pt]
    \multicolumn{10}{l}{\textbf{\textit{Full-parameter merge}}} \\
    Average                                  & \rowavg{50.4}{55.7}{60.7}{59.7}{60.2}{84.6}{32.4}{65.9} \\
    Task Arithmetic (TA)\textsuperscript{\dag}    & \rowavg{54.9}{61.6}{63.4}{64.7}{63.9}{88.3}{41.0}{71.8} \\
    TIES                                     & \rowavg{53.2}{58.5}{61.9}{61.2}{62.2}{88.7}{42.3}{65.9} \\
    DARE-TA                                  & \rowavg{54.4}{61.9}{63.1}{64.6}{64.2}{88.6}{41.7}{72.1} \\
    SCE                                      & \rowavg{50.1}{55.1}{61.5}{60.3}{61.1}{87.5}{37.1}{66.3} \\
    \addlinespace[2pt]
    \multicolumn{10}{l}{\textbf{\textit{LoRA merge}}} \\
    Average                                  & \rowavg{50.5}{54.8}{61.3}{60.6}{61.1}{85.8}{32.7}{65.8} \\
    Task Arithmetic (TA)\textsuperscript{\dag}    & \rowavg{54.3}{59.6}{61.9}{63.4}{63.2}{89.0}{41.9}{68.6} \\
    TIES                                     & \rowavg{49.1}{56.0}{61.0}{60.6}{61.0}{85.8}{32.6}{65.6} \\
    DARE-TA                                  & \rowavg{51.1}{58.4}{61.6}{62.6}{62.2}{87.7}{36.2}{67.5} \\
    Concat                                   & \rowavg{53.2}{58.1}{61.9}{63.6}{63.2}{88.9}{41.0}{68.6} \\
    SVD                                      & \rowavg{53.4}{58.2}{61.9}{63.2}{63.0}{88.7}{41.2}{67.5} \\
    \midrule
    \multicolumn{10}{c}{\textit{Qwen3-8B (non-thinking)}} \\
    \midrule
Base & \rowavg{19.0}{14.6}{46.0}{43.0}{45.9}{83.3}{25.9}{58.6} \\
+ Per-domain RL (Full)                   & \rowavg{39.2}{39.9}{52.5}{50.9}{50.9}{88.9}{42.8}{64.9} \\
    + Per-domain RL (LoRA)                   & \rowavg{38.1}{39.6}{52.4}{50.9}{52.3}{89.0}{42.0}{63.5} \\
    \addlinespace[2pt]
    \multicolumn{10}{l}{\textbf{\textit{Full-parameter merge}}} \\
    Average                                  & \rowavg{21.6}{17.5}{49.7}{45.0}{46.1}{84.0}{26.5}{60.5} \\
    Task Arithmetic (TA)\textsuperscript{\dag}    & \rowavg{41.4}{43.3}{54.4}{50.9}{51.4}{87.6}{37.0}{63.5} \\
    TIES                                     & \rowavg{27.1}{26.4}{52.4}{47.8}{48.7}{88.1}{37.8}{63.7} \\
    DARE-TA                                  & \rowavg{41.9}{43.1}{55.5}{52.4}{51.0}{88.0}{37.8}{63.5} \\
    SCE                                      & \rowavg{22.6}{19.3}{50.1}{46.3}{47.6}{86.4}{33.8}{62.3} \\
    \addlinespace[2pt]
    \multicolumn{10}{l}{\textbf{\textit{LoRA merge}}} \\
    Average                                  & \rowavg{22.4}{19.3}{49.6}{45.5}{47.2}{85.0}{28.7}{60.7} \\
    Task Arithmetic (TA)\textsuperscript{\dag}    & \rowavg{37.1}{40.4}{54.0}{50.0}{51.0}{88.2}{37.5}{63.8} \\
    TIES                                     & \rowavg{22.6}{19.3}{50.1}{44.9}{47.6}{84.7}{28.9}{61.7} \\
    DARE-TA                                  & \rowavg{27.9}{23.8}{52.5}{48.1}{49.7}{86.9}{34.6}{63.1} \\
    Concat                                   & \rowavg{34.4}{36.2}{54.5}{50.5}{51.1}{88.1}{37.3}{63.7} \\
    SVD                                      & \rowavg{34.0}{34.6}{54.3}{49.5}{51.1}{88.1}{36.9}{63.7} \\
    \bottomrule
    \end{tabular}
    }
    \end{center}
    \end{table*}

\section{Discussion and Limitations}
\label{app: discussion}

\paragraph{LoRA as a source of experts.}
LoRA provides a second, cheaper way to obtain the same five experts in this study.
Every adapter uses rank \(r = 32\) on all linear layers, and the resulting experts remain within \(1.4\) points of their full-parameter counterparts, close enough that both sets can be fused under identical conditions.
This lets us repeat the merge comparison in the low-rank regime, and the LoRA merges reproduce the domain profile of their full-parameter counterparts, with performance about one point lower on average.
The design answers one question cleanly---whether the parameterization of the experts changes what a merge retains---but leaves aside a second advantage of LoRA that fusion never calls on.
Because adapters are small and can be attached at serving time, a deployment can hold all \(N\) of them at once and route each query to the expert required by its domain~\citep{mindlab2026macaronv1preview}, and an adapter can in principle transfer a domain's update to another backbone rather than only to the one it was trained on~\citep{xia2025cross}.
Both routes sidestep the interference that Section~\ref{subsec: domain_interaction} traces to shared update directions, at the price of keeping \(N\) sets of parameters and committing to expert selection for every query.
They are therefore complements to the paradigms compared here rather than substitutes for them, and the domain relations we measure indicate which domains are worth keeping apart in the first place.

\paragraph{MOPD beyond its standard form.}
Of the three paradigms, MOPD is the most recent, and we study its standard form, the reverse KL to a domain-specific teacher computed over the states visited by the student's own rollouts~\citep{lu2025onpolicydistillation}.
This form is sufficiently well characterized for comparison with Merge and Mix RL, and it also explains the ceiling reported in Section~\ref{subsec: main_results}: the objective asks the student to match teacher behavior on its own samples, so it provides no signal for surpassing its teachers.
Recent work targets this limitation by extrapolating past the teacher with a reward signal~\citep{yang2026learning}, generalizing beyond a weaker teacher through policy shift~\citep{feng2026weak}, or revising the recipe from an analysis of what on-policy distillation transfers~\citep{li2026rethinking}.
Whether such variants lift the ceiling under multi-domain fusion remains unclear: the student must reconcile multiple teachers rather than follow a single one, and the domains themselves interact as Section~\ref{sec: analysis} shows.
The setup assembled here provides a direct testbed for studying such multi-domain extensions under controlled conditions, an avenue we leave to future work.

\end{document}